\documentclass[runningheads]{llncs}
\usepackage{graphicx}
\usepackage[ruled,vlined]{algorithm2e}
\usepackage{amsmath}
\usepackage{multirow}
\usepackage{amsfonts}

\newcommand{\x}{\textbf{x}}
\newcommand{\y}{\textbf{y}}
\newcommand{\X}{\textbf{X}}
\newcommand{\Y}{\textbf{Y}}
\newcommand{\z}{\textbf{z}}
\newcommand{\Z}{\textbf{Z}}
\newcommand{\G}{\textbf{G}}
\newcommand{\D}{\textbf{D}}

\begin{document}
\title{Task-Aware Active Learning for Endoscopic Image Analysis}
%
%
\author{Shrawan Kumar Thapa\inst{2} \and
Pranav Poudel\inst{3} \and
Binod Bhattarai\inst{1}\thanks{Corresponding author} \and 
Danail Stoyanov \inst{1}} 

\authorrunning{Thapa et al.}
%
\institute{
University College London, UK \and
NAAMII, Nepal \and
IOE Pulchowk Campus, Nepal \\
\email{\{b.bhattarai,danail.stoyanov\}@ucl.ac.uk}}
\maketitle              
\begin{abstract}
Semantic segmentation of polyps and depth estimation are two important research problems in endoscopic image analysis. One of the main obstacles to conduct research on these research problems is lack of annotated data. Endoscopic annotations necessitate the specialist knowledge of expert endoscopists and due to this, it can be difficult to organise, expensive and time consuming. To address this problem, we investigate an active learning paradigm to reduce the number of training examples by selecting the most discriminative and diverse unlabelled examples for the task taken into consideration. Most of the existing active learning pipelines are task-agnostic in nature and are often sub-optimal to the end task. In this paper, we propose a novel task-aware active learning pipeline and applied for two important tasks in endoscopic image analysis: semantic segmentation and depth estimation. We compared our method with the competitive baselines. From the experimental results, we observe a substantial improvement over the compared baselines. Codes are available at \url{https://github.com/thetna/endo-active-learn}.
\end{abstract}

\section{Introduction}
\label{sec:intro}
Polyp segmentation~\cite{brandao2017fully,ali2021deep} and depth estimation~\cite{rau2019implicit} are two fundamental research problems in endoscopic image analysis. Automatic polyp segmentation can help in the early diagnosis, detection and treatment of colorectal disease by supporting endoscopists with computer assisted detection and characterization systems. Meanwhile depth estimation is essential for 3D surface reconstruction of the endoluminal environment and for creating systems to map the colon shape and ensure that the full organ and all tissue surfaces are appropriately examined. Both capabilities are needed to advance the toolkit available to endoscopists and enable standardization of adenoma detection rates and potentially link to future robotic systems and automation~\cite{ahmad2021establishing}. 

\begin{figure}
    \centering
    \includegraphics[trim= 1.544 0cm 0cm 0.7cm, clip, width=0.24\textwidth]{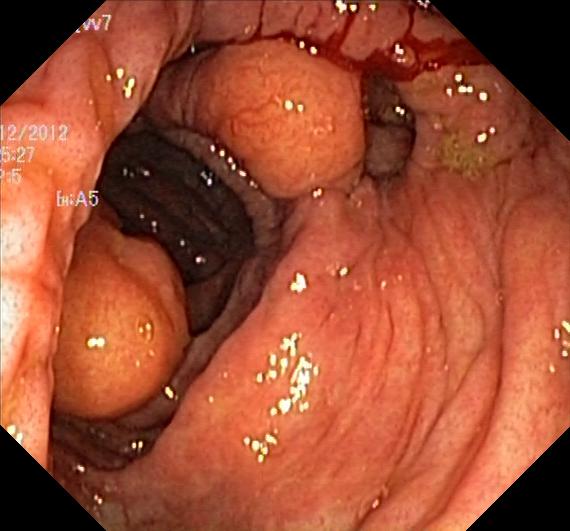}
    \includegraphics[trim= 1.544 0cm 0cm 0.7cm, clip, width=0.24\textwidth]{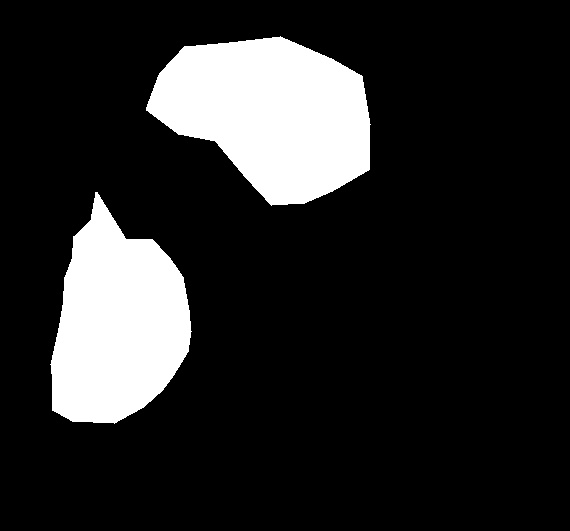}
    \includegraphics[trim= 1.544 0cm 0cm 0.7cm, clip, width=0.23\textwidth]{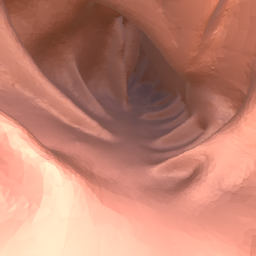}
    \includegraphics[trim= 1.544 0cm 0cm 0.7cm, clip, width=0.23\textwidth]{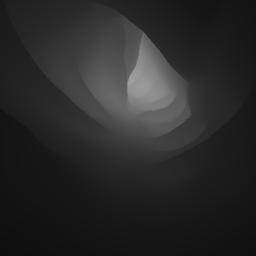}
    \caption{Two left  most images are from Kvasir-SEG~\cite{jha2020kvasir}, and the two right most images are from Colonoscopy Depth~\cite{rau2019implicit} data set.}
    \label{fig:intro}
\end{figure}

The efficacy of the deep networks to learn the parameters for both tasks, polyp segmentation and depth estimation, has been demonstrated, however, most solutions demand a large number of training examples. Annotating such a large volumes of endoscopic data is expensive, needs experts, and it is time-consuming, and for certain tasks like depth estimation, the ground truth is inherently not available. Figure~\ref{fig:intro} shows a pair of randomly selected images from Kvasir-SEG~\cite{jha2020kvasir} and Colonoscopy Depth~\cite{rau2019implicit} data sets. These are two 
important benchmarks publicly available for polyp segmentation and the depth estimation in endoscopy respectively. 

Active Learning (AL)~\cite{sener2018active,sinha2019variational,caramalau2021sequential} has shown a lot of promises to become a viable solution to sub-sample the data set by discarding redundant and less informative examples in computer vision. Nowadays, 
it is also slowly getting popular in Biomedical Image Analysis~\cite{budd2021survey,li2020imbalance}. NVIDIA's open source platform MONAI~\footnote{https:monai.io} has launched an intelligent interactive data annotation tool called MONAI Label. Workshop with the theme "interpretable and label-efficient learning"~\cite{cardoso2020interpretable} 
was organised in conjunction with MICCAI 2020. There are few works already published in premier venues in medical image analysis. For example, Yang et. al.~\cite{yang2017suggestive} proposed an AL framework for gland and 
lymph node segmentation based on class conditional uncertainty as a criterion to select unlabelled data.
Similarly,~\cite{mahapatra2018efficient} uses a conditional generative adversarial network to generate synthetic examples and estimate the uncertainties to select data to
query their labels. Relying only on uncertainty as a selecting criterion helps us to choose 
the examples from the region of the manifolds of the image where the model is less confident. However, it can not avoid selecting the redundant images
from the same manifold region, limiting the diversity. To address this issue, recently,
Shi et al.~\cite{shi2019active} designed an AL framework for skin lesion detection aiming to 
select both the difficult and the diverse examples. To this end, the paper 
proposed to do hashing on the image features computed in an unsupervised manner by applying Principal Component Analysis (PCA)~\cite{wold1987principal} and cluster the images into different bins. Next, the paper proposed to sample the images from each bin uniformly. This method addresses the issue of diversity that previous methods had. Nevertheless, the features computed in an unsupervised manner are not aware of the end task. Hence, the diverse examples on such sub-optimal features for the end task may not necessarily be diverse for the downstream task. As a result, the performance of the end task can be compromised.

To fulfil the criteria of task awareness and diversity in the selection mechanism, we proposed a novel framework for AL as shown in Figure~\ref{fig:proposed_method}. There are three main components in our pipeline: a Learner (A), a Sampler (B), and an oracle (C).  The learner is responsible for learning the parameters for a downstream task by using the examples selected by the sampler. The oracle's job is to query the labels of the examples selected by the learner. As the data selected by the sampler directly influences the learner's performance, we argue the need for linkage between these two components. Therefore, we propose to project all the examples from a data set on the learner's feature space  before applying the sampler. Afterwards, we apply K-Center Greedy Algorithm similar to that in~\cite{sener2018active} to select the core-set of the unlabelled data set. The number of unlabelled examples we select in any selection-stage depend upon the budget. The examples chosen by the samplers are sent to oracle to query the labels and repeat the cycle. 

We summarise our contributions in the following points. 

\begin{itemize}
    \item  We proposed a task aware active learning algorithm to select representative and diverse examples to annotate their labels.
    \item  We applied our method on two vital problems on endoscopic image analysis: polyp segmentation and depth estimation.
    \item We combined our method with uncertainty based AL method and observe improvement in the performance. 
\end{itemize}

\section{Method}
\label{sec:method}
\begin{figure}
    \centering
    \includegraphics[trim=2cm 0cm 0cm 0.7cm, clip, width=0.99\textwidth]{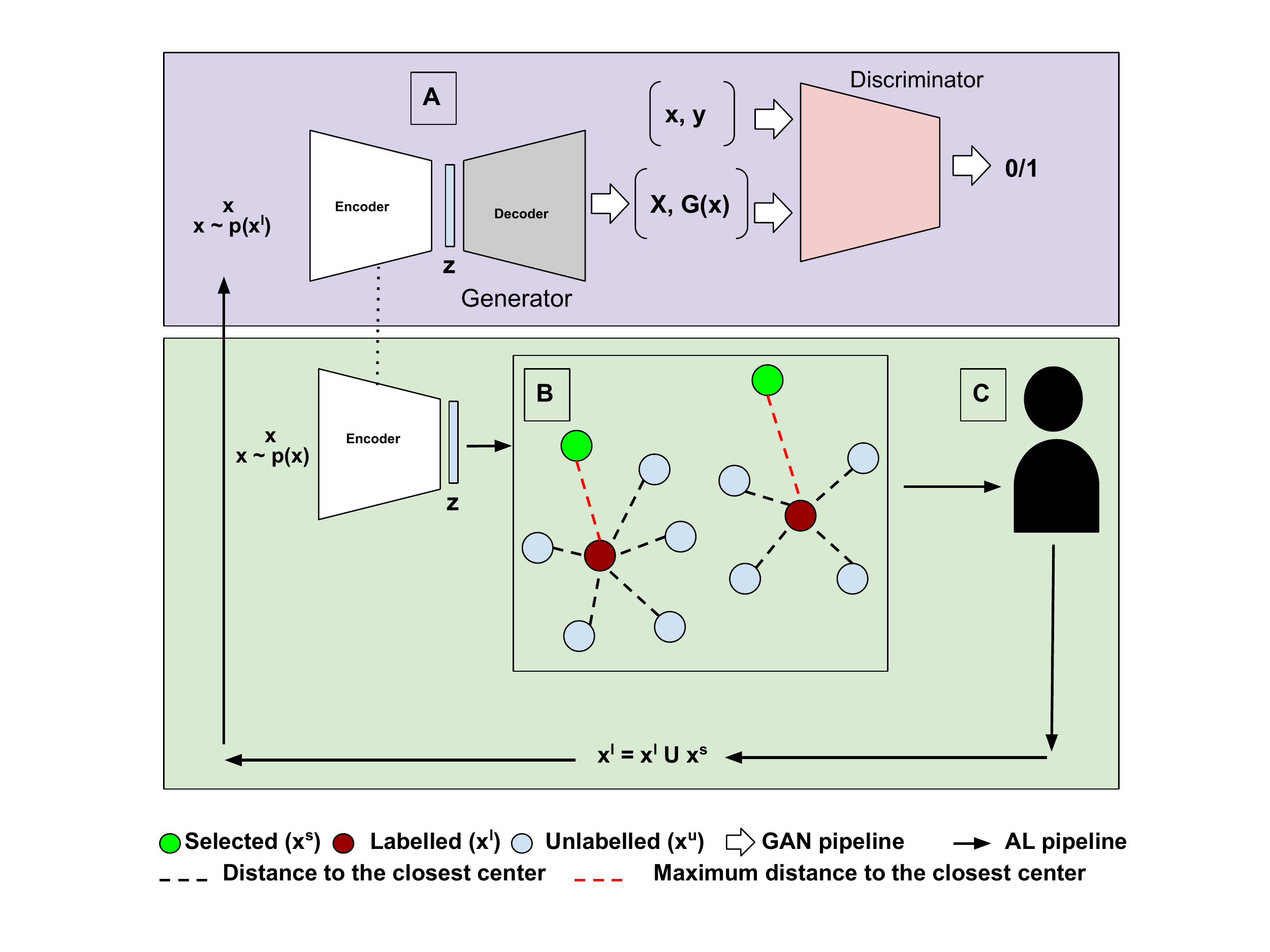}
    \vspace{-0.7cm}
    \caption{ Schematic diagram of the proposed Active Learning Pipeline.}
    \label{fig:proposed_method}
\end{figure}

Active Learning is an iterative process to select a subset of examples ($\X^s$) from a large pool of unlabelled set ($\X$) to query their labels~($\Y$). We label the examples 
($\x$, \y) $\subset$ ( $\X \times \Y$) incrementally and  add to a set of the labelled 
examples ($\X^l$). The labelled examples are used to train a network minimising the objective of the end task ($\mathcal{L}$). Equation~\ref{eq:2} summarises the Active Learning pipeline. Given any sampling function $\mathcal{A}$, the main goal of AL is to minimise the number of selection stages $n$ to reduce the number of examples to query their labels. 
In this paper, we  have considered two important tasks: semantic segmentation and depth estimation in endoscopic image analysis.

\begin{equation} \label{eq:2}
  \min_n \min_{\mathcal{L}} \mathcal{A}(\mathcal{L}(\x,\y;\theta)|\X^s_0\hspace{-1pt}\subset\hspace{-1pt} \cdots \hspace{-1pt}\subset\hspace{-1pt} \X^s_n\hspace{-1pt}\subset \hspace{-1pt} \X).
\end{equation}
To begin with annotation, we select the first batch $\X^s_0$ randomly, where subscript $0$ denotes the first selection stage and superscript $s$ indicates a selected set of examples to query their labels.  Once oracle queries their labels, we add those examples to the pool of labelled examples $\X^l = \{\X^s_0~U~\emptyset \}$. These labelled examples act as seed annotations to guide the next selection stages. Figure~\ref{fig:proposed_method} depicts the proposed method. There are three major components in the pipeline A) Learner, B) Sampler, and C) Oracle.  We discuss these in detail below. 
\\

\noindent \textbf{Learner (A):}
The role of a learner in the AL is to learn the parameters for a downstream task from the labelled set of examples.
In our case, we are tackling polyp segmentation and depth estimation, which are two key research problems in endoscopic image analysis. 
We choose pix2pix~\cite{zhu2017unpaired} to implement the learner. This architecture is well explored for semantic segmentation and depth estimation.  For example, Rau et al.~\cite{rau2019implicit} obtained state-of-the-art performance on Colonoscopy Depth data set. Similar architectures are widely being used to tackle the semantic segmentation problems~\cite{brandao2017fully}. 
The learner consists of a generator (\G) and a discriminator (\D). Generator is an 
encoder-decoder architecture while discriminator is a CNN network with convolutional layers. Suppose, $\x$ represents an image with its corresponding ground-truth label $\y$ from
labelled set $\X^l$. Depending on the problem taken into consideration, either 
semantic segmentation or depth map estimation, $\y$ will be a ground truth semantic
mask or a depth map. When we feed in $\x$ on the generator, the encoder projects
the image into a low dimensional vector, $\z$. And, the decoder reconstructs $\z$ back to the output \G(\x). We make a positive pair of ($\x, \y$) and ($\x, \G(\x)$) makes a negative pair. We feed these pairs to the discriminator to minimise the objective given in Equation~\ref{eqn:pix2pix_loss}.  Both the generator and the discriminator learn the parameters competing with each other. Once we learn the parameters of GAN from the available labelled examples, we move to component B of the pipeline, which is a sampler.
\begin{equation}
\mathcal{L}(\G, \D) =  \mathbb{E}_{(\x, \y)} [\log(\textbf{D}(\x, \y)]  + \mathbb{E}_{(\x, \G(\x)))} [\log(1 - \textbf{\D}(\x,  \G(\x))] 
\label{eqn:pix2pix_loss}
\end{equation}
\noindent \textbf{Sampler (B)}: In this stage, we use the parameters of the 
encoder learned in stage A to project both the labelled examples and  unlabelled 
images into a latent space, \Z. Since we train the learner to minimize the objective of the downstream task, the latent representations of the 
images are optimal for the downstream task. In Figure~\ref{fig:proposed_method} block B, each node 
represents an image and node's feature is initialised with the latent features extracted from the learner's encoder. 
We propose to select a core set~\cite{sener2018active} of the data 
set based on this latent space in order to query their labels. To compute the core-set, we employ K-Center-Greedy 
algorithm on this latent space as shown in Algorithm~\ref{algo:kcenter}.  Referring to the Figure~\ref{fig:proposed_method} 
block B, sky-blue nodes are unlabelled examples, red nodes are the labelled examples, and the green nodes are the selected examples to query their 
labels in the current stage. Please note, green nodes were sky-blue before selection. 
To select these green nodes, we compute the euclidean distance between the 
labelled examples  with every unlabelled examples. 
The edges between the nodes represent the euclidean distances. The length is proportional to the magnitude of the distances.
Higher the euclidean distance, longer is the edge. We create a territory of the nearest examples for each red nodes and select the farthest node amongst these ones as shown in the Figure. Examples with least euclidean distances are likely to be 
duplicates to the selected examples and does not provide a significant extra information to the downstream task. 
Hence, this selection strategy helps us to obtain a  subset of representative examples of the data set by discarding the examples which are less likely to add any new information for end task. 
As the core-set we obtain are aware of the downstream task, we called our approach as \emph{Task-Aware Coreset (TA-Coreset)}. The number of examples that we select in a stage depends upon the budget size, \textbf{b}. $\X^s$ makes the selected examples set.   

\noindent \textbf{Oracle (C)}: We send the selected set, $\X^s$ to the Oracle to query their labels. After retrieving their label, we append these examples along with their labels
to the set of labelled example ($\X^l =  \X^l \cup \X^s$) and empty the selected set ($\X^s= \emptyset$). With this, we repeat another cycle till our budget limit.

\SetKwInput{KwInput}{Input}                
\SetKwInput{KwOutput}{Output} 
\begin{algorithm}[H]
\caption{k-Center-Greedy}
\DontPrintSemicolon
\KwInput{ latent representation of data $\mathbf{Z}$, existing pool $\mathbf{s}^0$ and a budget $b$}
Initialize $\mathbf{s} = \mathbf{s}^0$ \\
\Repeat{$|\mathbf{s}| = b + |\mathbf{s}^0|$}{
    $u = \arg\max_{i\in[n]\setminus s} \min_{j \in s}\Delta(\mathbf{z}_i, \mathbf{z}_j)$\\
    $\mathbf{s} = \mathbf{s} \cup \{u\}$
}
\Return $\mathbf{s} \setminus \mathbf{s}^0$ 
\label{algo:kcenter}
\end{algorithm}

\section{Experiments}
\label{sec:experiments}
\noindent \textbf{Implementation Details:}
We trained pix2pix GAN as our task model for both depth estimation and polyps segmentation. We used Adam optimizer ($\beta1=0.5, \beta2=0.999$)  with a learning rate of $2 \x 10^{-4}$ for both generator and discriminator and used 200 as $L_1$ loss weight as in \cite{rau2019implicit}. We trained the model for 100 epochs in each cycle with batch size of 20 for depth estimation and batch size of 8 for polyps segmentation. Unlike in \cite{rau2019implicit}, we used Instance Normalisation at each convolutional layer of both discriminator and generator. 

For AL experimentation, we initialized our labelled pool with randomly selected 1310 examples for depth estimation, while keeping the sampling budget size of 655. 
In case of polyps segmentation, we initialized with randomly selected 100 examples keeping the same sampling budget size for following acquisition stages. 
\\
\noindent \textbf{Data sets:} We performed experiments on
Colonoscopy Depth~\cite{rau2019implicit} and Kvasir-SEG~\cite{jha2020kvasir}. Colonosocpy Depth is a synthetic data set consists of 16,380 images  with their depth annotations. We divided the dataset into train, validation and test set in the ratio of 8:1:1, giving us 13104 training examples. Similarly, Kvasir-SEG consists of 1000 real images with polyp masks. We used 900 of them for training, and rest for validation. We reported our performance on a smaller test dataset provided by the same project consisting of 196 images.   
\\ 
\noindent \textbf{Baselines:}
We have compared our method with a wide range of competitive baselines. \textbf{Random} is the most commonly used technique to sub-sample the training examples. We also applied Coreset~\cite{sener2018active} on low-dimensional feature representation computed in an unsupervised manner. We applied Principal Component Analysis (PCA)~\cite{wold1987principal} and compressed the images to the dimension of 512 which is equal to that of the latent representations in our method.
\textbf{Uncertainty}~\cite{joshi2009multi} is another sampling technique to find the most informative examples.  Finally, we also compared our performance with \textbf{VAAL}~\cite{sinha2019variational} which is one of the current state-of-the-art task-agnostic active learning method. 
\\
\noindent \textbf{Evaluation metrics:} We report the performance of depth estimation in Root Mean Square Error (RMSE). Whilst, we evaluate the performance of semantic segmentation in mean Intersection Over Union (mIOU).

\begin{figure}
    \centering
    \includegraphics[trim= 1.544 0cm 0cm 0.9cm, clip, width=0.70\textwidth]{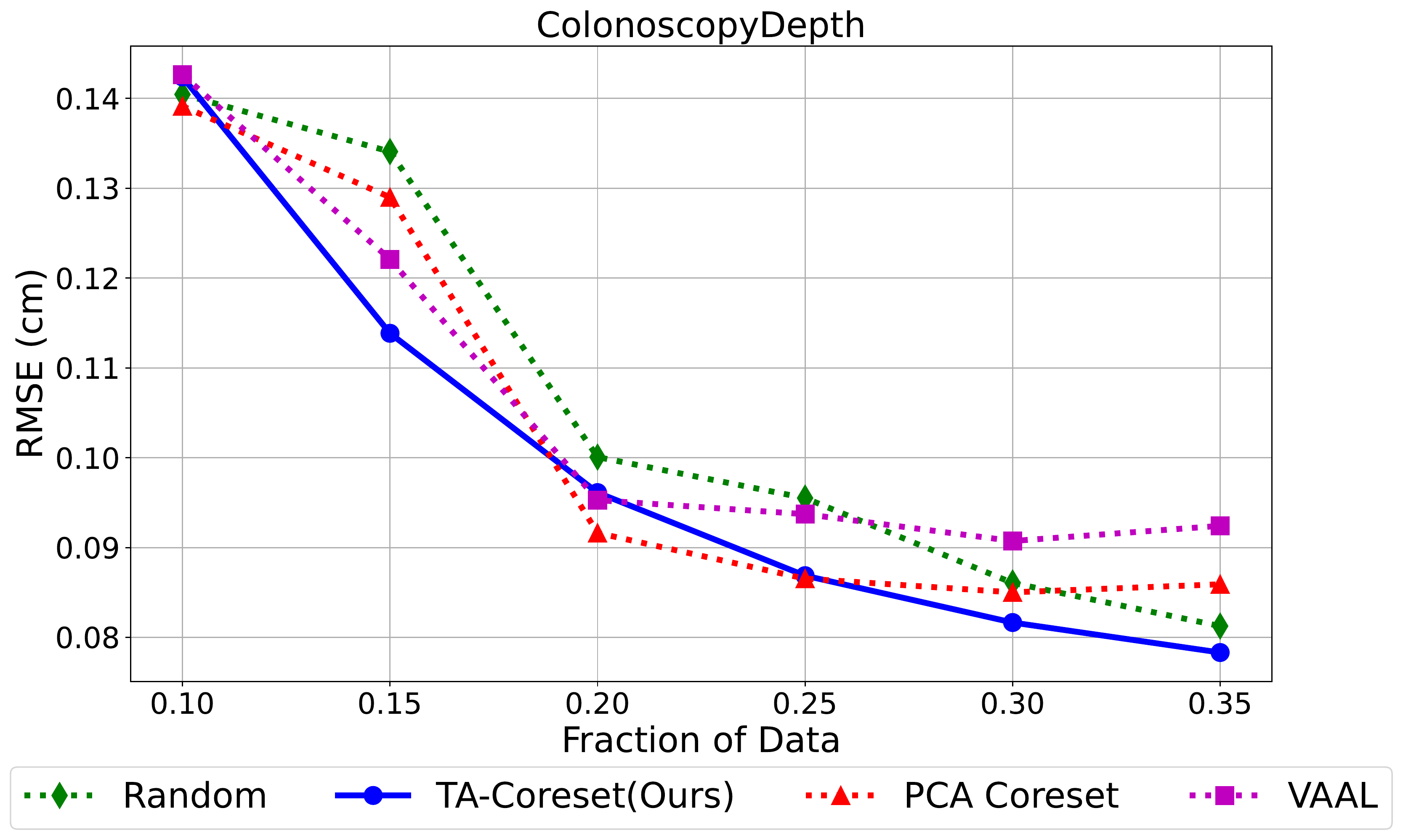}
    \caption{ Performance comparison on Colonoscopy Depth data set. 
    }
    \label{fig:perf_depth_estimation}
\end{figure}

\subsection{Depth Estimation:} Figure~\ref{fig:perf_depth_estimation} shows  the  performance  comparisons 
between  the  baselines  and  the proposed method. In the figure, the x-axis shows the fraction of total available
data used to train the learner and the y-axis shows their corresponding RMSE
reported  in  cm.  From  the graph,  we  can  see  that  all  the  AL  frameworks outperform the random sampling technique. Among the AL methods, the
proposed method outperforms the others in most cases. PCA-coreset
and VAAL are two task agnostic methods employing the sampling method on latent
representations  of  the  images.  The  performance  of  these  two  methods 
lags  behind  our pipeline. This gap in performance highlights  the  need  of  making  the  representations  of  the images task-aware. The performance of Random with 25\% of examples is almost equivalent to the performance of TA-Coreset with 20\% examples(or 30\% with 25\%). This is 20\% reduction in the training examples.

\subsection{Semantic Segmentation:} We also employed our method for polyp segmentation. Table~\ref{tab:seg_perf_table} (upper block) summarises the performance comparison with the baselines.
The first 100 examples in the experiments were randomly selected. In the table, we can observe that the performance of the random sampling is least compared to every AL methods. Among the AL methods, our TA-Coreset outperforms the rest in the every selection stages. Superior performance of our method compared to the task agnostic methods PCA-Coreset and VAAL validates our hypothesis. The performance of Random with 300 examples is almost equivalent to the performance of TA-Coreset with just 200 examples and so on.  
\setlength{\tabcolsep}{12pt}
\begin{table}
    \centering
    \begin{tabular}{|r|r|r|r|r|r|}
        \hline
        \multirow{2}{*}{\textbf{Method}} & \multicolumn{5}{c|}{\textbf{Mean IOU}} \\ \cline{2-6}
                                & \textbf{100} & \textbf{200} & \textbf{300} & \textbf{400} & \textbf{500} \\
        \hline
        Random & 57.1 &  59.2 & 61.1 & 65.3 & 70.7 \\
        \hline
        VAAL~\cite{sinha2019variational} & 57.1 & 61.2 & 63.9 & 68.0 & 72.6 \\
        \hline
        PCA-Coreset & 57.1 & 58.2 & 62.2 & 66.8 & 70.2 \\
        \hline
        TA-Coreset (ours) & 57.1 & \textbf{61.8} &  \textbf{66.9} & \textbf{71.2} & \textbf{74.5} \\
        \hline
        \hline
        Uncertainty + &  57.1 & 60.0 &  63.0 & 65.8 & 72.6 \\
        PCA~\cite{shi2019active} & ~ & ~ & ~ & ~ & ~ \\ 
        \hline
        Uncertainty +&  57.1 & \textbf{62.8} & \textbf{67.1} & \textbf{71.5} & \textbf{76.1}\\
        TA-Coreset & ~ & ~ & ~ & ~ & ~ \\ 
        \hline

    \end{tabular}
    
    \caption{Performance Comparison on Kvasir-SEG Dataset. 
    }
    \label{tab:seg_perf_table}
\end{table}

\noindent \textbf{Combining TA-Coreset with Uncertainty:} 
In order to validate whether our method is complementary to another category of AL methods relying on uncertainties, we combine our method with that of Joshi et al.~\cite{joshi2009multi}. This is one of the most competitive baselines to select the informative examples and uses the difference of the best and the second best  confidence score (BvSB) approach \cite{joshi2009multi}. Less the difference in class prediction, more uncertain the model is about the example. We first sample $N_u$ uncertain examples using uncertainty score given by BvSB. Then, we sample $N_d$ diverse examples using TA-Coreset. The numbers $N_u$ and $N_d$ are controlled by a parameter $\gamma$ such that $N_u = \gamma * N$ and $N_d = (1 - \gamma) * N$, where N is the total number of examples at a selection stage. Tab.~\ref{tab:seg_perf_table} (lower block) shows the performance comparison with~\cite{shi2019active}. From the table, we can see that our method is complementary to existing methods based on uncertainties and also outperforms one of the state-of-the-art methods.

\section{Qualitative Evaluations}
\begin{figure}
    \vspace{-1cm}
    \centering
    \includegraphics[trim= 0cm 0cm 0cm 1.1cm, clip, width=0.32\textwidth]{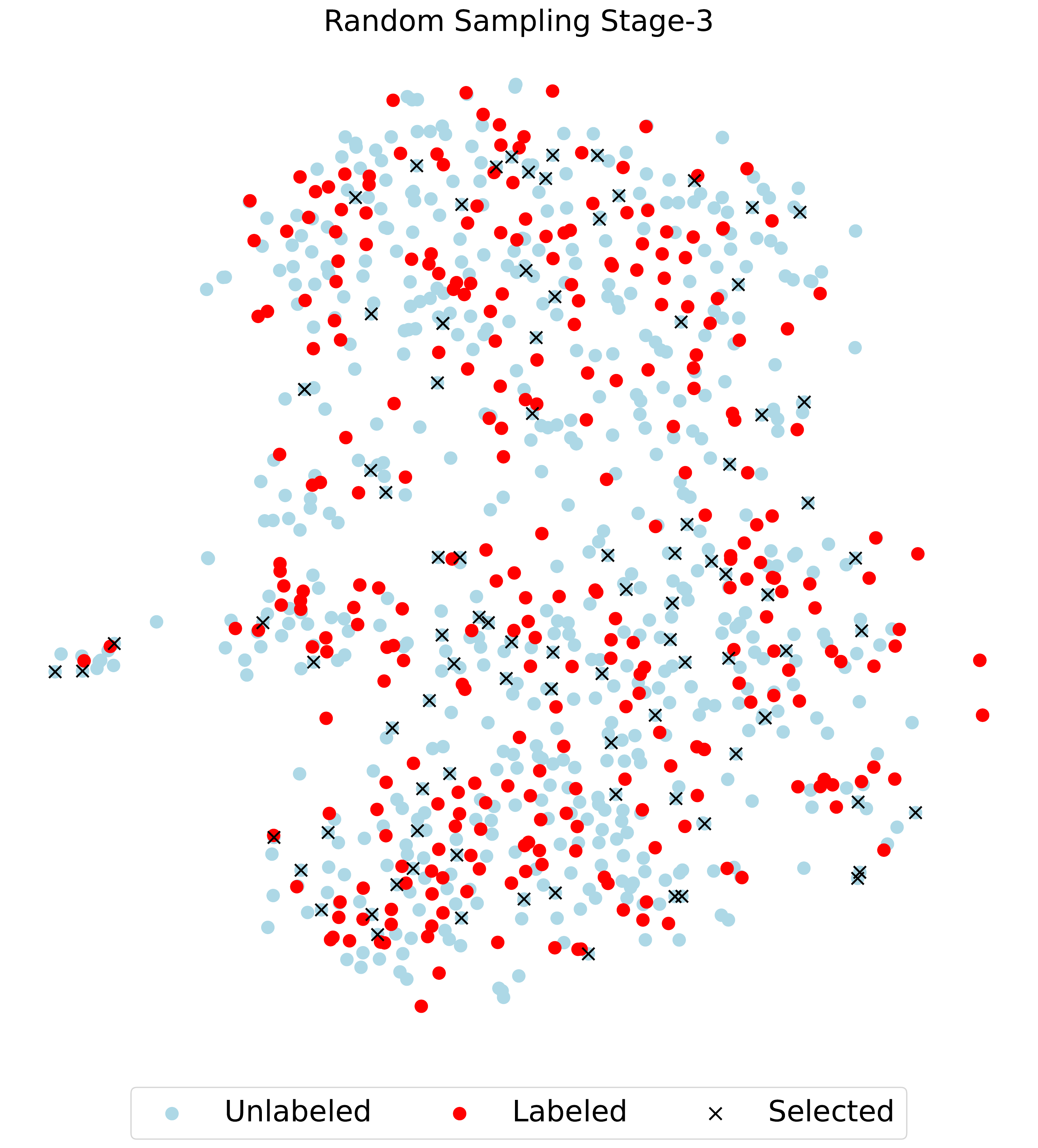}
    \includegraphics[trim= 0cm 0cm 0cm 1.1cm, clip, width=0.32\textwidth]{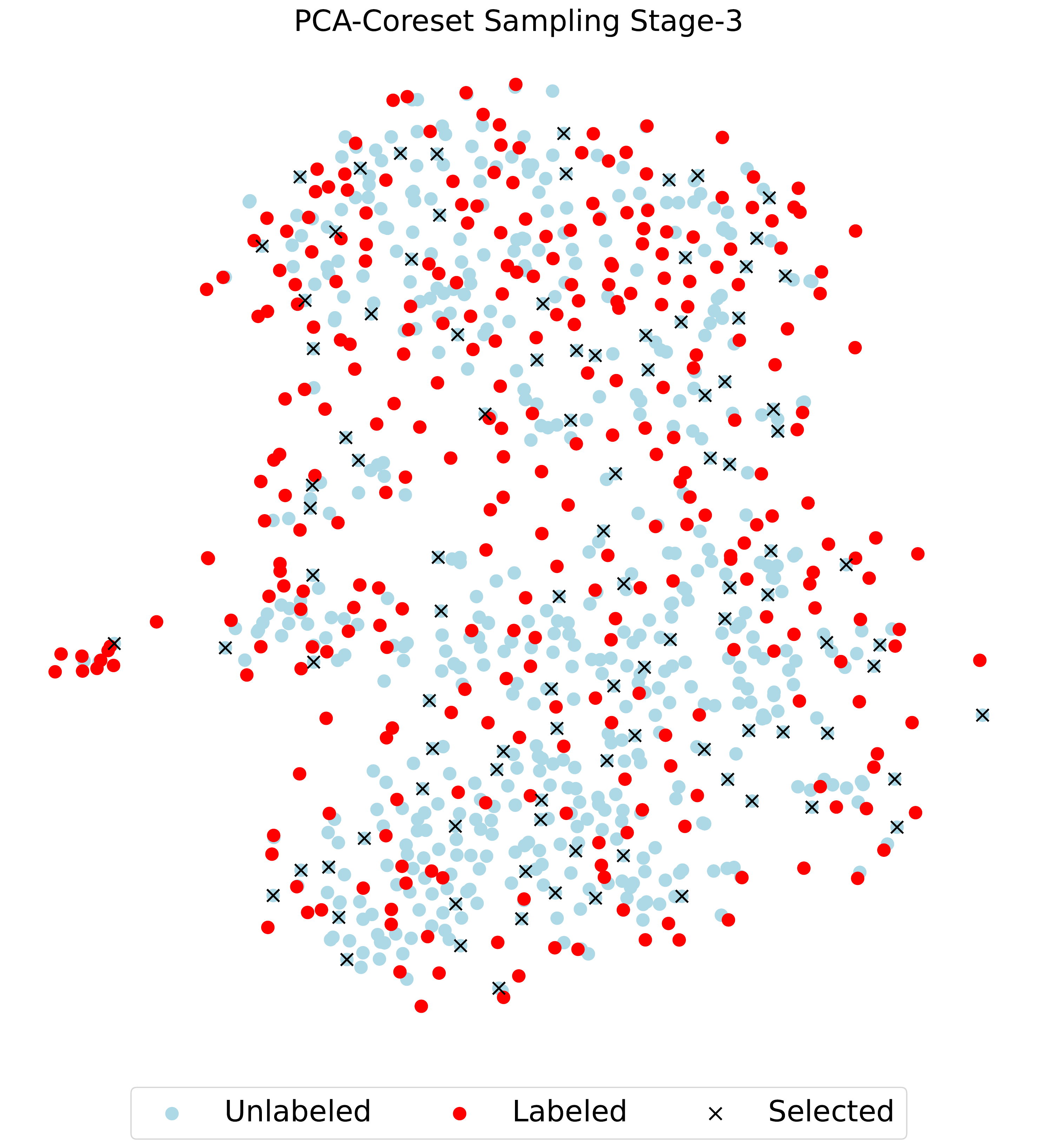}
    \includegraphics[trim= 0cm 0cm 0cm 1.1cm, clip, width=0.32\textwidth]{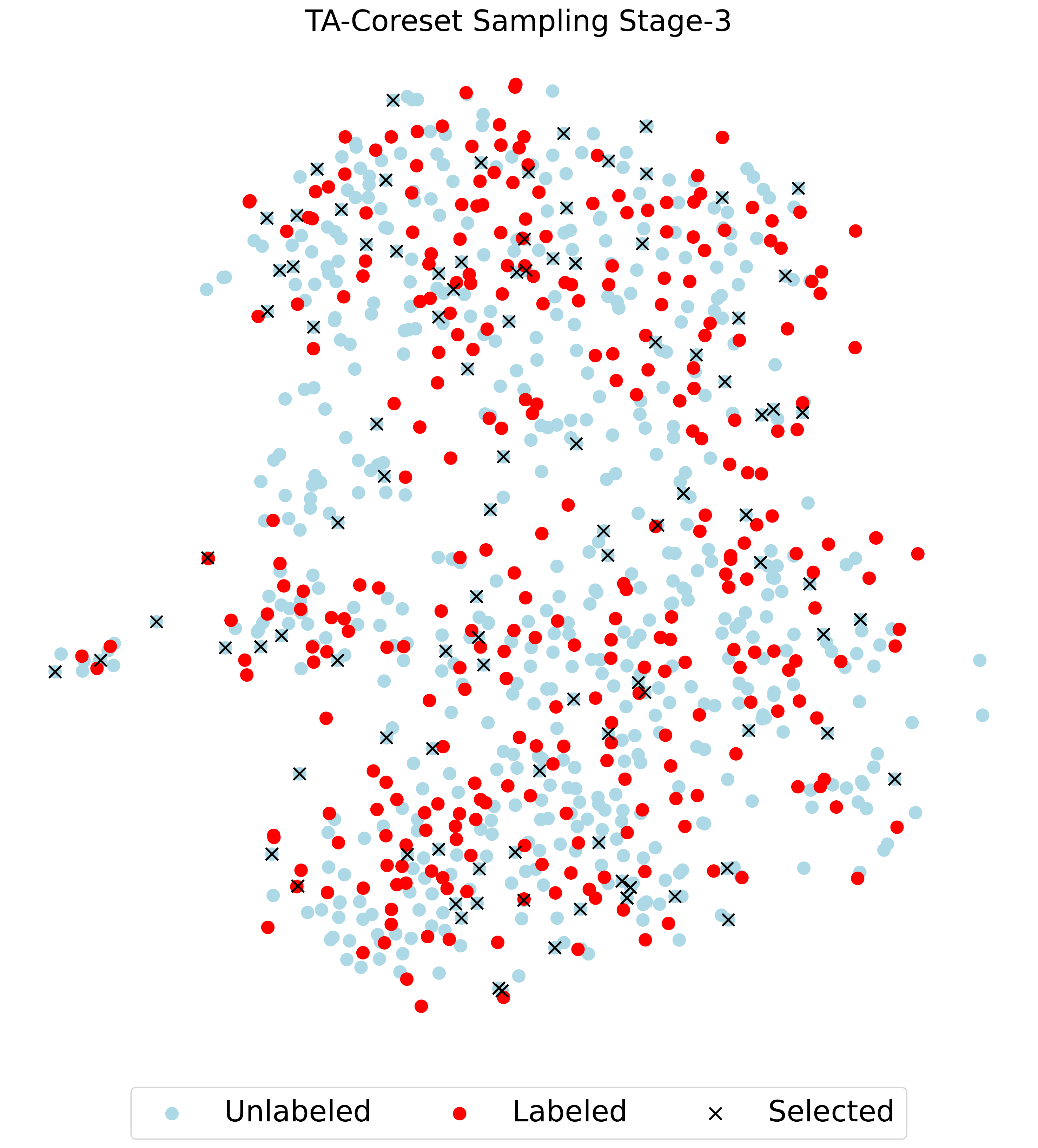}
    \caption{T-SNE plots showing the comparison of selection of unlabelled examples at the fourth selection stage on Kvasir-SEG. Left, middle, and right plots show selection by Random, PCA-Coreset, and TA-Coreset, respectively (Zoom in for the better view).}
    \label{fig:qual_tsne_plot}
\end{figure}

In the Figure~\ref{fig:qual_tsne_plot}, we summarize the selection behaviour of different sampling techniques in 
Kvasir-SEG data set with help of the tSNE~\cite{van2008visualizing} plots. In the diagram, the grey dots are unlabelled examples,
red dots labelled examples in previous selection stage, and the cross represents the examples selected in the
current-stage for annotation. 
From these plots, random (left) selects the examples uniformly throughout the manifold. Similarly, PCA-coreset (middle)
also covers the whole manifold. While, TA-Coreset (right) covers manifold as a whole and 
also concentrate on some of the regions middle of the image manifold. These depict that sampling in task-aware latent representation is more important than selecting diverse examples on the task-agnostic image feature space.

\section{Conclusions}
In this paper, we present a novel task-aware active learning framework for endoscopic image analysis. We employed the proposed method on polyp segmentation and depth estimation.  We observe a superior performance from the extensive experiments compared to the multiple competitive baselines, validating the proposed idea.  Moreover, the generalisation of our method on two different tasks make it a generic method and exhibits a high potential to be equally effective in other tasks. We combined our method with the existing AL method and observed their performance improvement. And it
demonstrates a complementary nature of our method.

\section{Acknowledgement}
This project is funded by the EndoMapper project by Horizon 2020 FET (GA 863146). For the purpose of open access, the author 
has applied a CC BY public copyright licence to any author accepted manuscript version arising from this submission.

\bibliographystyle{splncs04}
\bibliography{refs}
\section{Appendix}
\begin{figure}
    \centering
    \includegraphics[trim= 1.544 2cm 0cm 1.544cm, clip, width=0.75\textwidth]{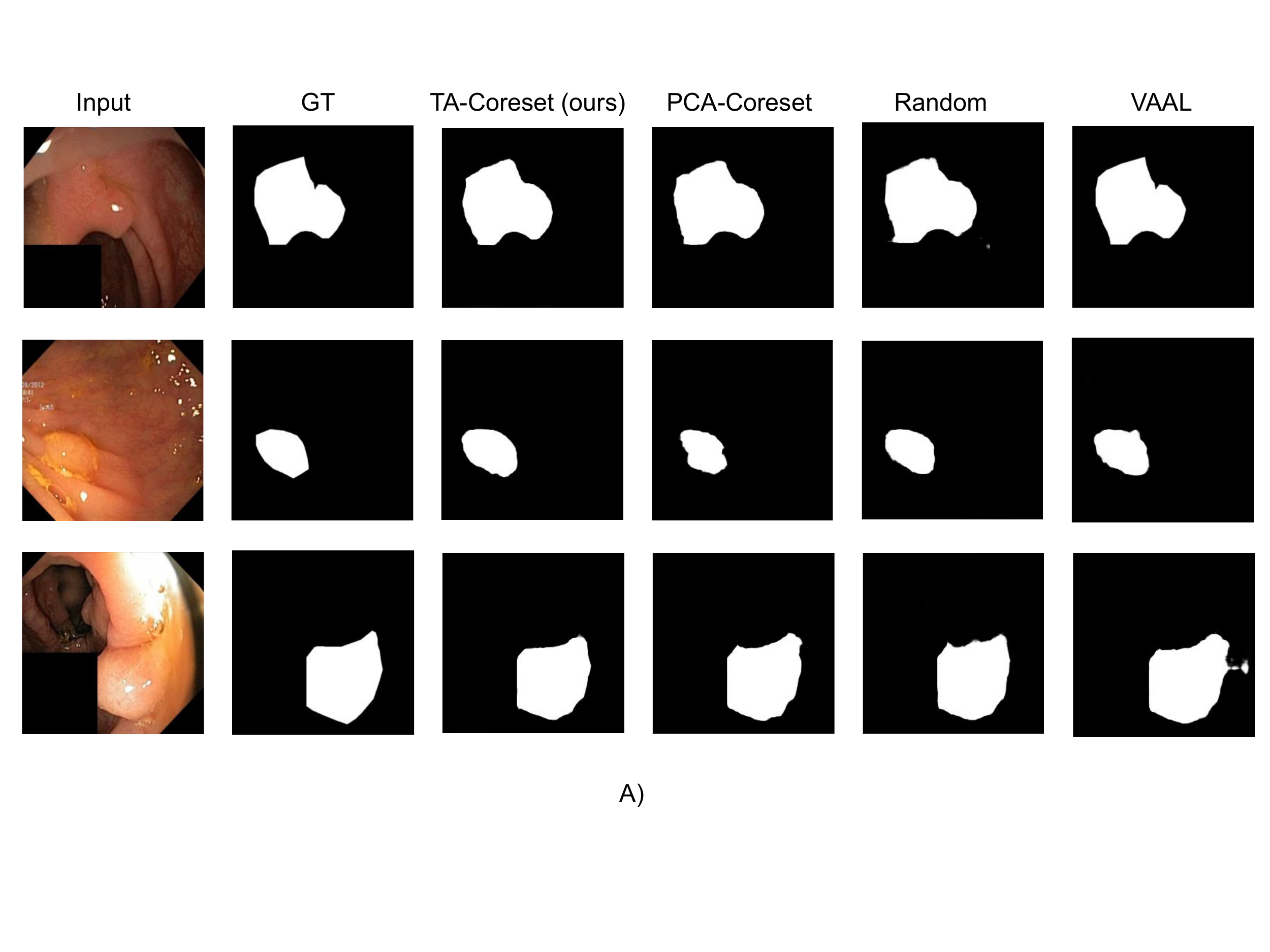}
    \includegraphics[trim= 1.544 3cm 0cm 1.544cm, clip, width=0.75\textwidth]{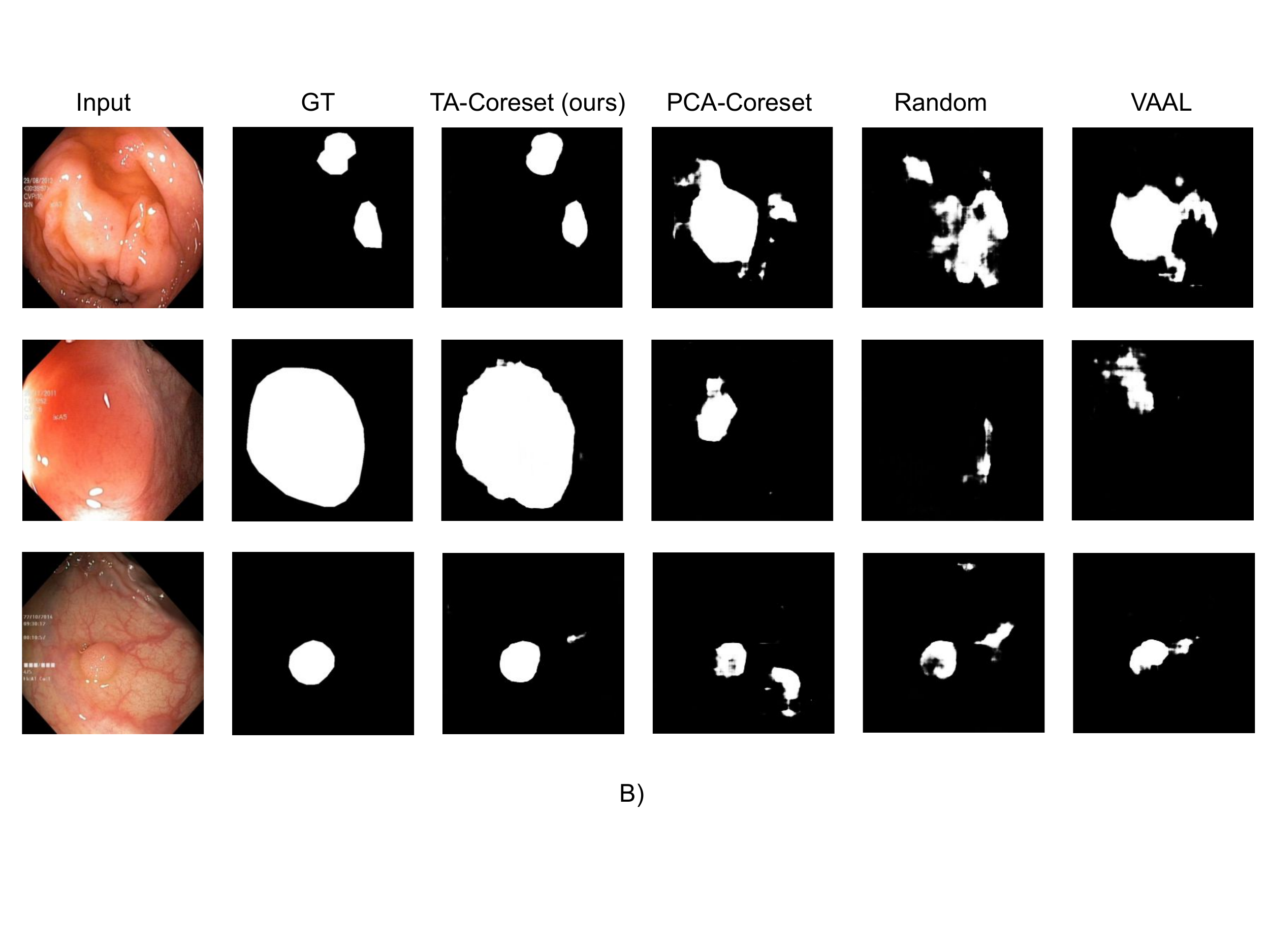}
    
    \caption{Qualitative comparisons on Kvasir-SEG Dataset\cite{jha2020kvasir}: A) Examples with Similar Performance B) Examples where our method performs best}
    
    \label{fig:qualitative_seg}
\end{figure}

\begin{figure}
    \centering
    \vspace{-0cm}
    \includegraphics[trim= 1.544 2cm 0cm 1.544cm, clip, width=0.75\textwidth]{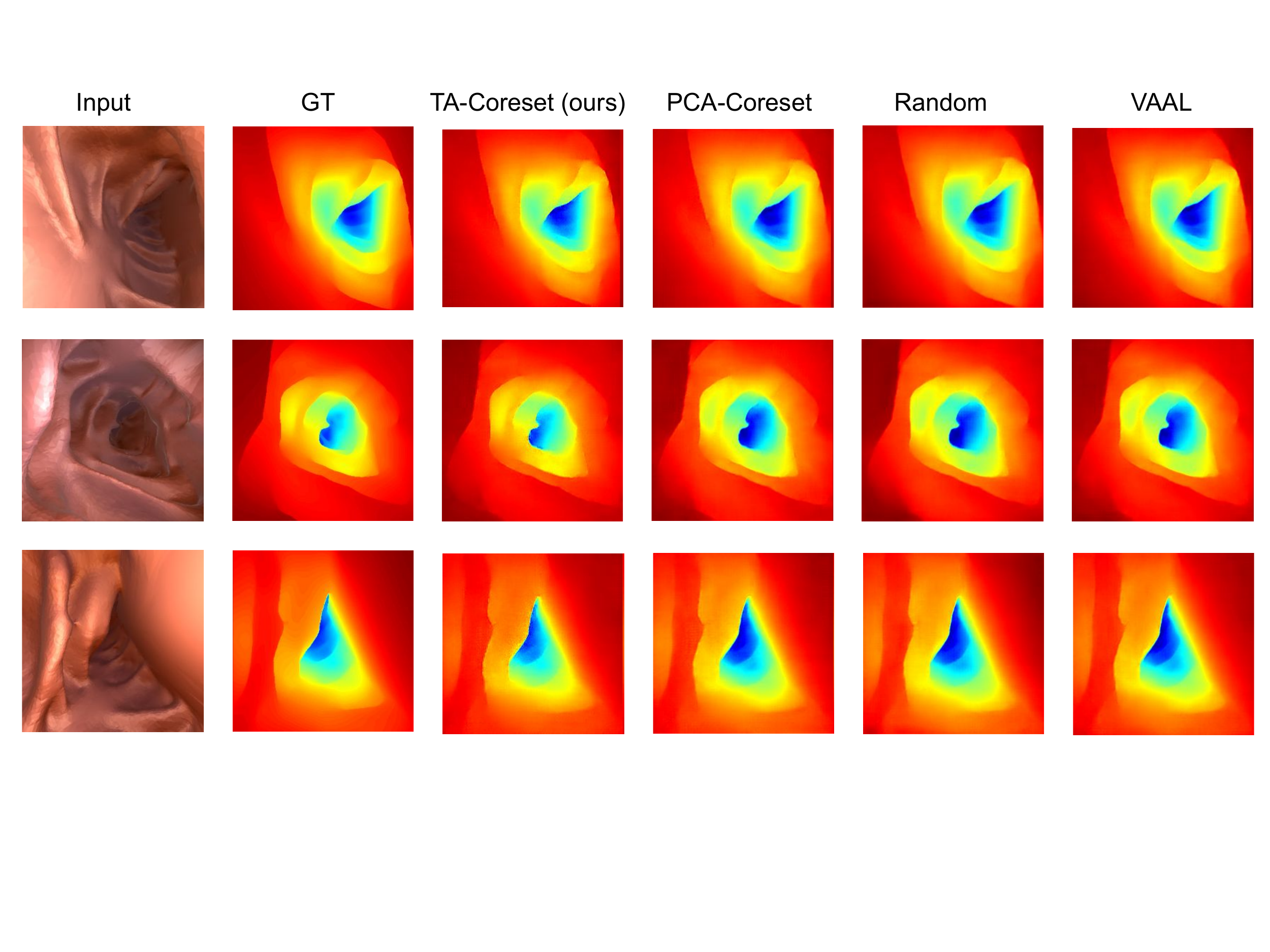}
    
    \caption{Qualitative comparisons on Colonoscopy Depth Dataset}
    
    \label{fig:qualitative_depth}
\end{figure}

Figure~\ref{fig:qualitative_seg} and ~\ref{fig:qualitative_depth} show the qualitative comparison of the models trained on data annotated at the \emph{third} selection stage. From these comparisons, we can observe that the proposed method is able to select the examples from the manifolds where the model is uncertain, whereas the baselines fail (1 B). 

\section{Combining Uncertainty and TA-Coreset: Ablation}
\begin{figure}
    \centering
    \includegraphics[trim= 1.544 0cm 0cm 0.9cm, clip, width=0.70\textwidth]{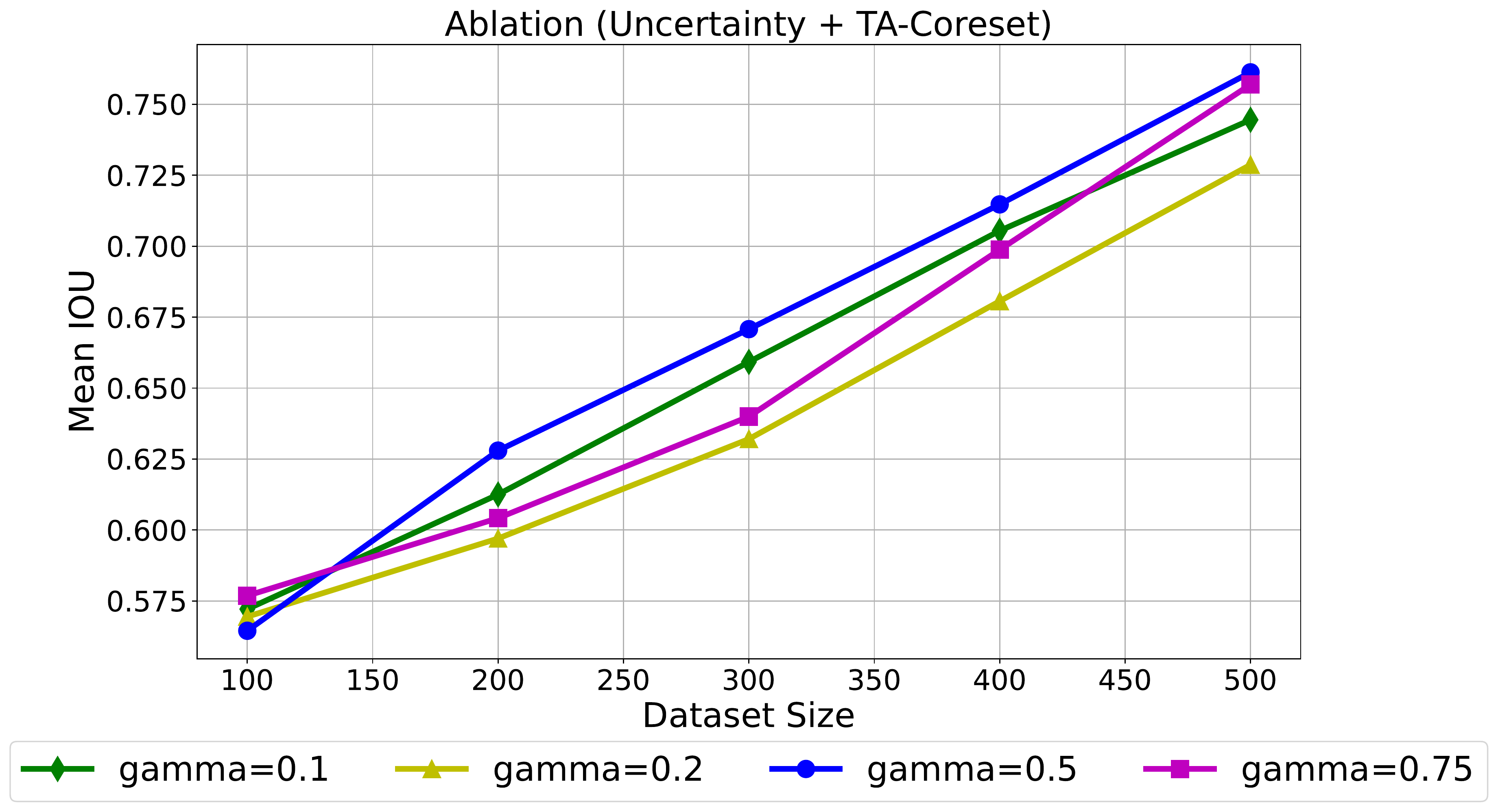}
    \caption{ Ablation study on different combinations of 
    Uncertainty and TA-Coreset}
    \label{fig:ablation_unc_coreset}
\end{figure}

We performed experiments on combining samples from Uncertainty based acquisition function and TA-Coreset. We experimented with different values of $\gamma$, which corresponds to fraction of uncertain examples to sample from total budget. The test results on Kvasir-SEG \cite{jha2020kvasir} is shown in figure \ref{fig:ablation_unc_coreset}. We found out that equally balancing uncertainty and diversity($\gamma=0.5$) while sampling performed best. The setting closer to TA-Coreset ($\gamma=0.1$) showed similar trend to TA-Coreset itself. We see similar trend to its counterpart in case of heavy uncertainty sampling ($\gamma=0.75$) as well.

\end{document}